\documentclass[conference]{IEEEtran}

\usepackage[utf8]{inputenc}

\usepackage{amsmath}
\usepackage{amssymb}
\usepackage{graphicx}
\usepackage{subcaption}
\usepackage[colorlinks=true, allcolors=black]{hyperref}
\usepackage[backend=biber,maxnames=2,url=false,
  style=numeric-comp,isbn=false,giveninits=true,eprint=false,
  sorting=none,doi=false,date=year]{biblatex}
\bibliography{main.bbl}

\DeclareMathOperator{\Xc}{\mathcal{X}}
\DeclareMathOperator{\Yc}{\mathcal{Y}}
\DeclareMathOperator{\Zc}{\mathcal{Z}}
\DeclareMathOperator{\Dc}{\mathcal{D}}
\DeclareMathOperator{\Rb}{\mathbb{R}}

\usepackage[normalem]{ulem} %strike out

\title{\huge On Manifold Learning in Plato's Cave:\\
    Remarks on Manifold Learning and Physical Phenomena}

\author{Roy R. Lederman and Bogdan Toader\\Yale University}

\begin{document}

\maketitle

\begin{abstract}
Many techniques in machine learning attempt explicitly or implicitly to infer a low-dimensional manifold structure of an underlying physical phenomenon from measurements without an explicit model of the 
phenomenon or the measurement apparatus.
This paper presents a cautionary tale regarding the discrepancy between the geometry
of measurements and the geometry of the underlying phenomenon in a benign setting. 
The deformation in the metric illustrated in this paper is mathematically 
straightforward and unavoidable in the general case, and it is only one of several 
similar effects. While this is not always problematic, we provide an example 
of an arguably standard and harmless data processing procedure where this 
effect leads to an incorrect answer to a seemingly simple question.
Although we focus on manifold learning, these issues apply broadly to dimensionality 
reduction and unsupervised learning.
\end{abstract}

\section{Background} 
\label{sec:main}

The abundance of data in many applications in recent years allows scientists to 
sidestep the need for parametric models and discover the structure of 
underlying phenomena directly from some form of intrinsic geometry in the 
measurements. Such concepts frequently appear in unsupervised learning, manifold 
learning, non-parametric statistics and, more broadly, machine learning.
Often, a scientist may have in mind a concept of the ``natural'' geometry 
or parametrization of the phenomenon; in other cases, they may implicitly assume 
that only one such objective geometry exists even if they do not know what it is.
This paper aims to illustrate the difference between the structure of {\em observed} data and some notion of {\em natural} or {\em unique objective} structure. 
To this end, we offer a concrete example with an obvious underlying natural 
geometry (up to symmetries) and demonstrate the existence of discrepancies 
between the data and the natural variables, even in this benign setting.

In our example, described more formally below, a simplified instance of a physical 
phenomenon is represented by a rigid 3D model of a horse on a spinning table. The measurement device is a fixed camera that takes images of the object. 
The orientation angles of the horse are distributed uniformly.
Here, a natural variable is the angle at which the figure is oriented at the time of the measurement. 
A simple example of a scientific question is to find the mode of the  
distribution, which is intuitively the most prevalent orientation angle (we know that the correct answer is that the distribution is uniform and, therefore, we do not expect to find a clear mode).
Since this is meant to be a simplified, intuitive version of a generic problem 
with no obvious underlying model, we consider generic algorithms and 
forgo in advance image analysis and computer vision methods that use of the 
special properties of images and the specific rotating motion of the object.

This benign task yields results that we find surprising yet predictable.
The naive analysis discovers clear modes of the distribution, which are 
inconsistent with the true uniform distribution. In Appendix~\ref{sec:two cameras}, 
we demonstrate that these modes are not invariant to the measurement modality.

In our discussion, we explain the reasons for the experiment's results and 
refer the reader to existing work on special cases where the problem can be 
corrected. However, there is no method for correcting the problem in the general case.
We conclude by pointing to where care should be taken in defining the problem and using the output in downstream tasks.

We emphasize that this paper aims to highlight an omission that we 
observe in the practical use of manifold-related machine learning algorithms in applications. The purpose of this paper is not to advocate against these methods but rather to suggest that care should be taken in stating and interpreting their output.

\section{The problem}

\begin{figure}%[h]
    \centering
    \includegraphics[width=\linewidth]{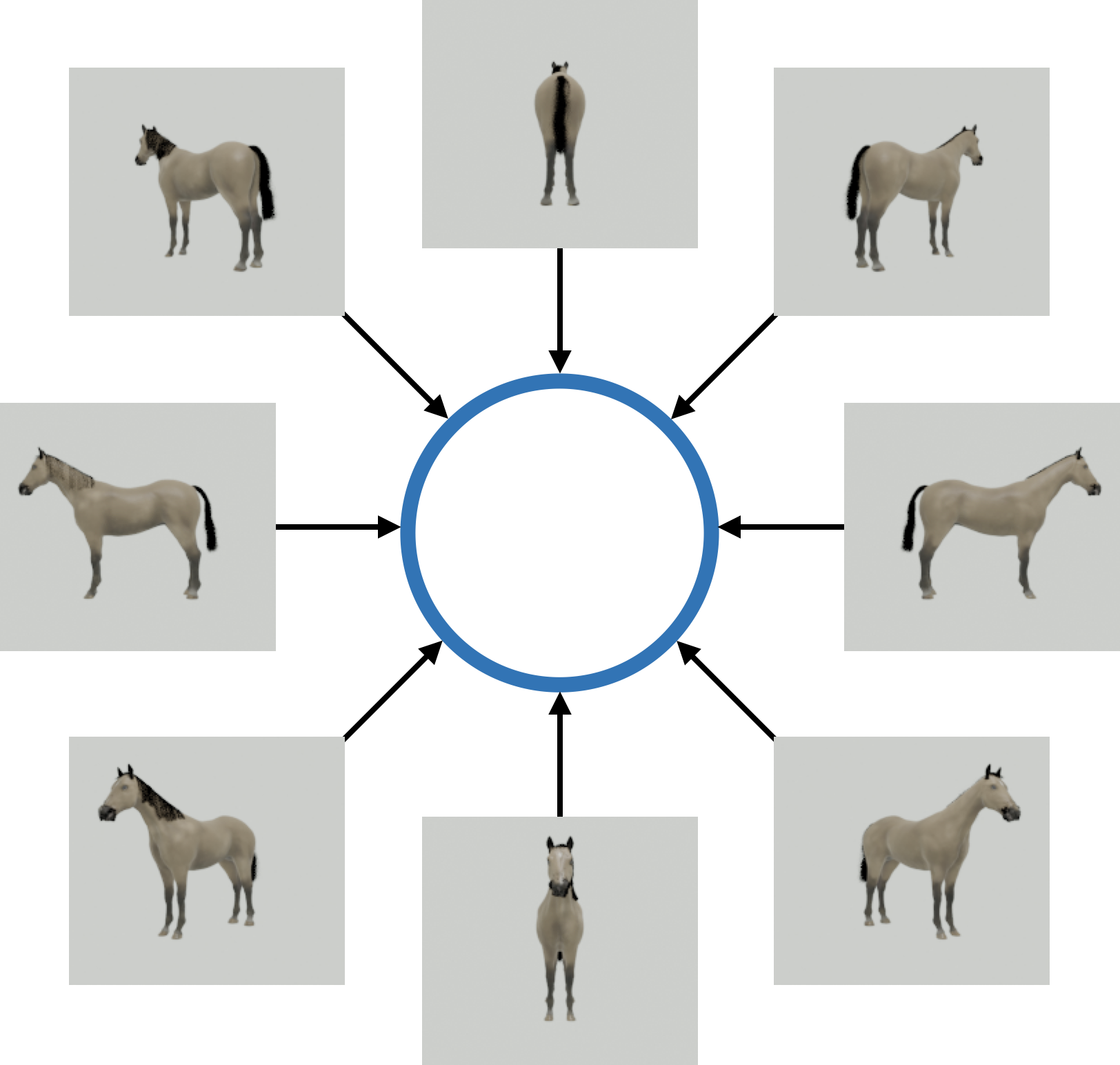}
    \caption{The phenomenon manifold $\Xc$ is a one-dimensional torus corresponding 
        to the in-plane orientation angle of a rigid object rotating around 
        the $z$-axis, and the measurement manifold $\Yc$ is the manifold of images of the 
        object as captured by a camera at a fixed location.}
    \label{fig:setting}
\end{figure}

The mathematical setting of the experiment is simple: let $\Xc \subset \Rb^d$
and $\Yc \subset \Rb^D$ be two manifolds with $d \ll D$
and $f: \Xc \rightarrow \Yc$ be a diffeomorphism. 
We refer to $\Xc$ and $\Yc$ as the {\em phenomenon manifold} and the 
{\em measurement manifold} respectively and to $f$ as the {\em measurement 
function}. 
In our simple experiment, the phenomenon manifold $\Xc$ is the one-dimensional torus
representing the orientation of the horse with respect to a fixed frame of
reference (independent of the camera), 
the measurement function $f$ outputs an image of the horse as captured by the 
camera, and the measurement manifold $\Yc$ is 
the manifold of images obtained by the camera.
In particular, a sample $x \in \Xc$ is the angle of the horse 
at a specific point in time, and the corresponding measurement $f(x)\in\Yc$ 
is the image of the horse at the same point in time. In a typical setting, we 
are given a large set of measurements $\left\{ y_i \right\}_{i=1}^n \subset \Yc$
of a set of samples $\{x_i\}_{i=1}^n$ drawn from a distribution $\Dc$ on $\Xc$. 
Here, we take the distribution $\Dc$ of the orientation angles of the horse to 
be uniform, which would be unknown in an actual experiment.
We only have access to the measurements $\{y_i\}_{i=1}^n$ (the images of the horse), 
which we assume to be noise-free for simplicity, and we are interested in uncovering 
the low-dimensional organization of the samples of angles $\{x_i\}_{i=1}^n$, 
for example, their empirical distribution on $\Xc$. The setting of this numerical
experiment is illustrated in Figure~\ref{fig:setting}.
For simplicity and concreteness, we apply common techniques to answer a 
simple question: what is the most dominant physical state? We know that the ground 
truth answer is, in this case, that there is no dominant state; the data are generated 
with uniform distribution over the orientation angles of the horse.

We follow a common practice of assuming a low-dimensional structure 
and apply a manifold learning algorithm. This produces the map $\rho : \Yc \to \Zc$, 
which yields a low-dimensional embedding of the measurements $\rho(y_i) \in \Zc$ for $i=1,\ldots,n$ 
and $\Zc \subset \Rb^s$ with $d \leq s \ll D$. In our experimental setting,
the low-dimensional assumption is clearly true: the orientation angles of the 
horse lie on the one-dimensional torus manifold, while the measurements are 
clearly high-dimensional (the number of pixels of each image). For simplicity,
in our numerical experiment, we use the diffusion maps algorithm, whose 
theoretical properties are well understood~\cite{lafon_diffusion_2004,coifman_diffusion_2006},
and we retain only the first two diffusion coordinates, a standard practice in 
this simple case.
The output we expect to see is an embedding of the one-dimensional torus 
in $\mathbb{R}^2$: a circle.

It is common in applications to apply a machine learning or manifold learning algorithm 
to the measurements $\{y_i\}_{i=1}^n$, and consider the low-dimensional embeddings $\{\rho(y_i)\}_{i=1}^n$ to be
a proxy for the geometry of the actual samples $\{x_i\}_{i=1}^n$; the 
potential effects of the measurement function $f$ are omitted. 
The aim of this manuscript is to demonstrate that even in the most benign setting, 
the measurements distort the physical problem in a way that can impact a seemingly 
straightforward analysis.

Many algorithms for manifold learning and visualization have been developed over 
the years and have been found useful in applications.
Often, these algorithms start with the pair-wise distances 
$\|y_i - y_j\|$ (in some norm), for $i,j=1,\ldots,n$, as a measure of (inverse) 
similarity, 
but diverge in their precise formulation of the problem. 
One of the notable departures from this approach is the use of the latent space 
estimated in the training of deep neural networks as the manifold embedding,
with the variational autoencoder (VAE)~\cite{kingma_auto-encoding_2014} being
one of a number of popular approaches.

The diffusion maps algorithm produces coordinates that are related 
to the geometry of the data through a diffusion operator on the data manifold. 
While there are technical nuances in the metric defined by diffusion maps 
(and other algorithms) and in retaining only two dimensions, this example 
is particularly benign, symmetric, and without boundary effects. 
Therefore, one expects the leading eigenvectors of the discretized diffusion operator to 
preserve the local geometry of the data (up to scaling). 
For a formal description of the diffusion maps algorithm and its properties, 
see~\cite{lafon_diffusion_2004,coifman_diffusion_2006}. 
One of the appealing properties of the diffusion maps algorithm is that it is 
(asymptotically) invariant to the local density of the data and captures only its
local geometry. 
This property and the algorithm's explicit relationship to the geometry of the 
data made it a good choice for our experiments. 

Indeed, a diffusion map of the points on $\Xc$ preserves the 
geometry and the uniform distribution (shown in Appendix~\ref{sec:ml on x}). 
However, our measurement function is not necessarily an isometry (even up to 
scaling), and therefore, it distorts the geometry and the local pair-wise distances. 

\begin{figure}%[h]
    \centering
    \begin{subfigure}{0.25\textwidth}
        \centering
        \includegraphics[width=\linewidth]{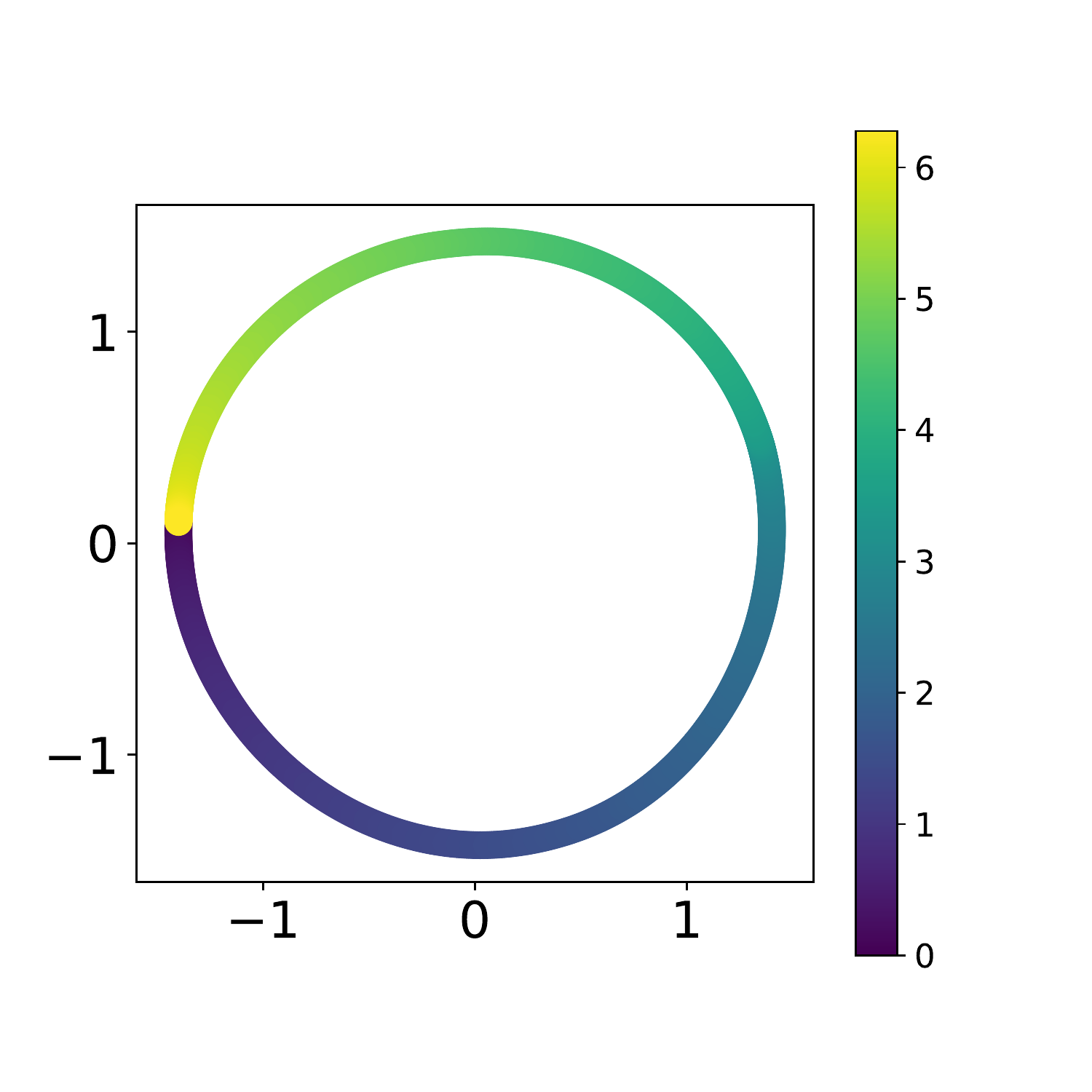}
        \caption{True angle}
        %\label{fig:sub1}
    \end{subfigure}%
    \begin{subfigure}{0.25\textwidth}
        \centering
        \includegraphics[width=\linewidth]{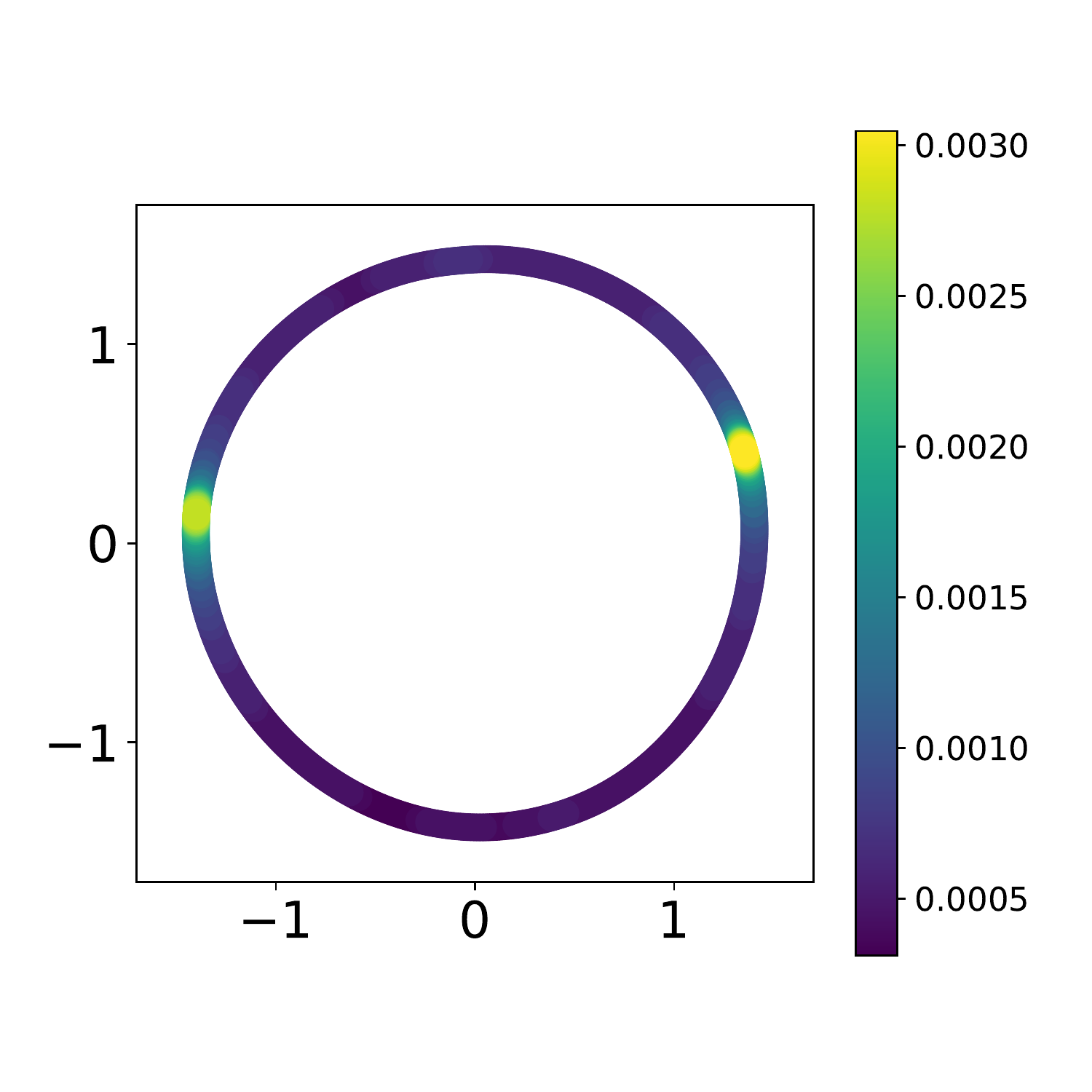}
        \caption{Local density}
        %\label{fig:sub1}
    \end{subfigure}
    \caption{Low-dimensional embedding of the images of the spinning horse.
        The coloring is given by the true orientation angle of the horse in
        panel (a) and the local density of points ($r=0.05$) in panel (b).}
    \label{fig:main result}
\end{figure}

The low-dimensional embedding obtained by applying the diffusion maps algorithm
to a dataset $\{y_i\}_{i=1}^n$ of size $n=1000$ and ambient dimension $D=108000$ ($180\times200$ size images with $3$ color channels)
in our experimental setting\footnote{
    The code and dataset to reproduce the numerical experiments described in this paper 
    can be downloaded from \url{https://github.com/bogdantoader/ManifoldLearningInPlatosCave}.
} 
is shown in Figure~\ref{fig:main result}.
Both panels show a scatter plot using the first two 
embedding coordinates given by the diffusion maps algorithm. 
The points in panel (a) are colored according to the true angle $x_i$ for $i=1,\ldots,n$. 
Visually, it appears that the algorithm reveals the correct topology and 
it organizes the images correctly by their angle. It is compelling to say that the embedding is a good approximation of the angles (up to shift).
However, taking a closer look at the distribution of the points in panel (b), 
we see that that their \textit{local density}\footnote{
    We define the local density at a point $z_i$ on the embedded manifold $\Zc$ 
    as the number of points $z_j$ in the ball centered at $z_i$ with radius
    $r$, for a given $r > 0$, normalized so that the densities sum up to one,
    and using the metric on $\Zc$. Since in these
    experiments, we used the diffusion maps algorithm with a two-dimensional
    latent space, the diffusion metric on $\Zc$ corresponds to the Euclidean 
    metric on $\mathbb{R}^2$.
} has not been preserved on the embedding manifold $\Zc$:
while the distribution $\Dc$ of the points $\{x_i\}_{i=1}^n$ on $\Xc$
is uniform (by construction!), the distribution of the 
embedded points $\{\rho(y_i)\}_{i=1}^n$ on $\Zc$ is not uniform.
Moreover, the distribution of the embedded points has two clear modes, 
with no indication that they are an artifact of the analysis.

An additional experiment showing how the distribution of the embedded points
varies when the viewing angle is changed is described in 
Appendix~\ref{sec:two cameras}, and the specific implementation of the 
diffusion maps algorithm that we used in our experiments is presented in
Appendix~\ref{sec:dm}.

\section{Discussion}
\label{sec:discussion}

In the previous section, we empirically showed how the distribution of the points on the 
embedding manifold $\Zc$ does not reflect the true distribution of the points 
on the phenomenon manifold $\Xc$: the distribution on $\Zc$ has two distinct 
modes, while the distribution on $\Xc$ is uniform. To see that this is a 
metric-related issue, it is worth examining the modes of the distribution on~$\Zc$.

\begin{figure}%[h]
    \centering
    \includegraphics[width=\linewidth]{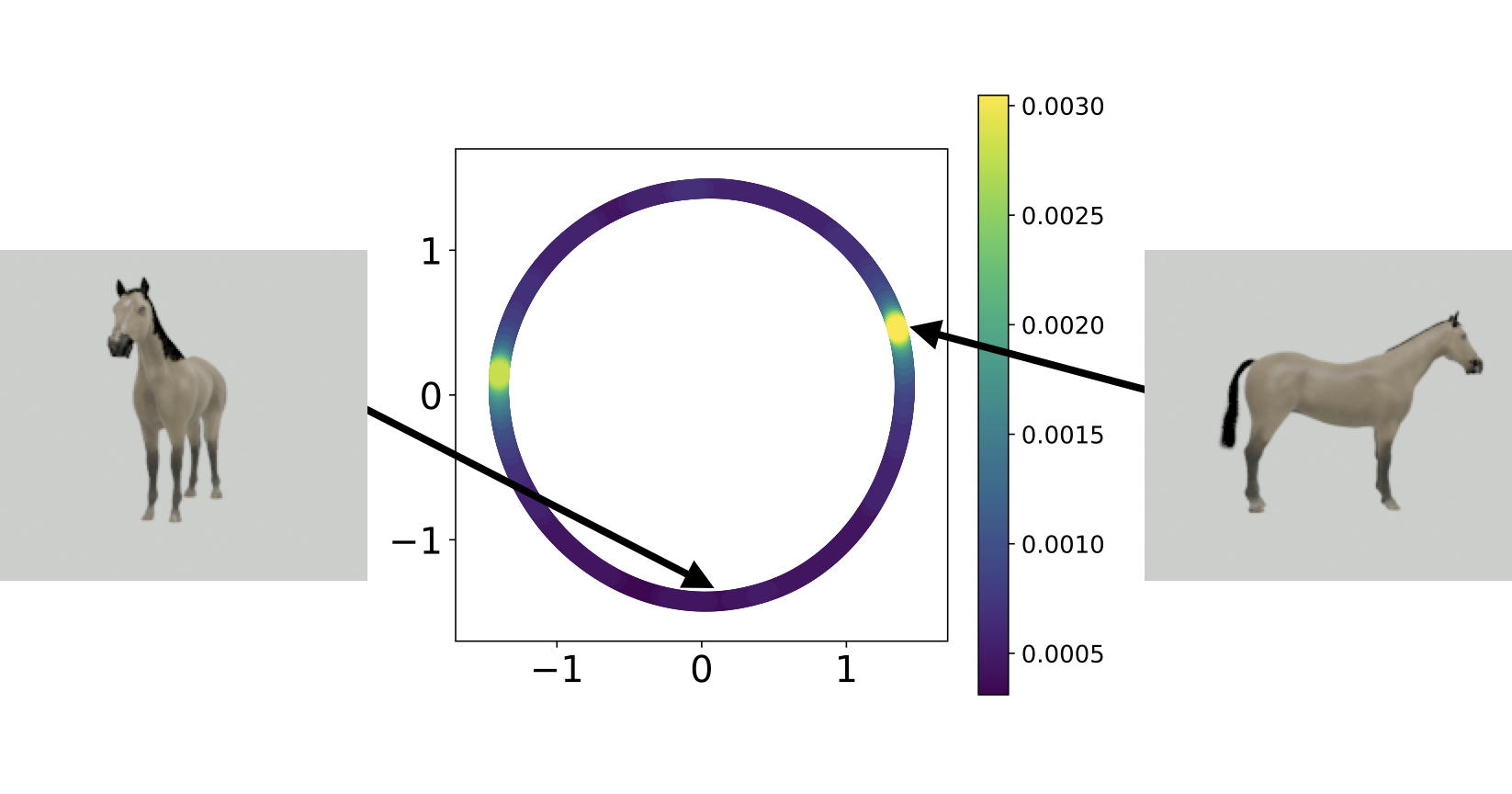}
    \caption{The low-dimensional embedding with example images corresponding to
        samples from the estimated distribution. The image to the left of the 
        embedding plot is chosen to be at a low local density in the 
        embedding, and the image to the right is chosen to be at the maximum 
        density point.}
    \label{fig:main result with horse}
\end{figure}

In Figure~\ref{fig:main result with horse}, we show measurements at a high
and a low-density point on $\Zc$. It is revealed that the high-density regions 
correspond to images where the three-dimensional object is perpendicular (or 
nearly perpendicular) to the viewing direction of the camera, while the 
low-density regions correspond to the object facing toward or away from the 
camera. This is because, according to our chosen metric on $\Yc$ (i.e., 
the Euclidean norm on the space of vectorized images), a small difference 
$\Delta x$ between two angles in $\Xc$ is not transformed to the same distance
in different regions of $\Xc$: two images of the object facing the camera that 
differ by $\Delta x$ have a larger Euclidean distance than two images of 
the object facing sideways that are separated by the same angle. 
The metric based on the measurements alone %chosen here 
does not account for the distortion introduced by 
the measurement function $f$ on the true metric on $\Xc$, namely the wrap-around 
distance on $[0,2\pi)$. 

The discrepancy between the metric on the phenomenon manifold, which is the metric
we want to recover, and the arguably arbitrary metric produced by the measurement 
modality can be corrected in some special cases. 
For example, when bursts of measurements around each point on $\Xc$ are available, 
one can use the Jacobian of the measurement function to define metrics 
that are invariant to the measurement modality 
(see, for example, \cite{singer_non-linear_2008,talmon_empirical_2013,peterfreund_local_2020,schwartz_intrinsic_2019,bertalan_transformations_2021}).
Such metrics might still not be the ``desired'' metrics we want to conceptualize, 
but they are ``Platonic'' in the sense that they are defined on the phenomenon 
manifold $\Xc$ and are invariant to the arbitrary measurement function.
Other works such as \cite{perraul-joncas_non-linear_2013,mcqueen_nearly_2016,little_estimation_2009,little_multiscale_2017,little_multiscale_2017} 
correct the metric distortion introduced by the embedding $\rho$ from the measurement 
manifold $\Yc$ to the embedded manifold $\Zc$; these works do not correct the discrepancy between the measurements and the metric on the phenomenon manifold. 

We emphasize that the problem illustrated here is not due to a failure of the diffusion maps or other algorithms;
the algorithm performs as expected and characterizes the {\em measurement manifold} very well.
However, the metric of this measured manifold is incompatible with the natural metric of in-plane
rotation angles.
As a result, we identify modes of the distribution in the measurement space, but these do not correspond to modes of the underlying distribution of angles.

We note that the problem discussed here is not unique to the diffusion maps algorithm or the setup we chose;
in fact, other algorithms are not as well-understood as diffusion maps, and applications are 
rarely as simple as our illustrative example.
Many modern algorithms add layers of complexity to the problem. For instance, deep learning approaches that generate 
latent variables, such as VAEs, are often combined with more 
standard manifold learning algorithms to obtain low-dimensional data representations.
In~\cite{cooley_novel_2022,chari_specious_2022}, the distortions introduced by popular 
algorithms like t-SNE and UMAP are analyzed in the context of single-cell genomics,
although the focus is on the discrepancy between the high dimension of the measurement
space and the very low dimension (2 or 3) of the embedding space, rather than
on the choice of metric.
While such algorithms provide valuable new insights into datasets, practitioners
should be aware that the results they generate, even when they perform as intended, 
may have a subtle relation to the ``Platonic'' physical reality.
These outputs should arguably mainly be used for visualization and
confirmed by other means.
Indeed some of the original work on popular non-linear dimensionality reduction
algorithms defines them as tools for visualization~\cite{van_der_maaten_visualizing_2008,mcinnes_umap_2018}.

\section{Conclusions}

This paper illustrates one of the discrepancies between the measured manifold and a perceived natural parametrization of the underlying phenomenon.
In addition, Appendix~\ref{sec:two cameras} demonstrates how this discrepancy depends on the measurement modality and how the measured manifold is not invariant to measurements.
The discrepancy presented here is by no means the only type of discrepancy; 
we defer the discussion of additional effects to future work.
While the existence of this discrepancy is a natural consequence of various mathematical formulations of manifold learning problems (with the exception of special cases where the metric can be corrected), it is occasionally omitted, which may lead to incorrect and inconsistent answers to seemingly simple scientific questions.
In the absence of a general solution to the problem, we suggest the following 
points to consider when using these methods.
\begin{itemize}
    \item A good rule of thumb is that manifold learning and dimensionality reduction can 
    provide (when they ``work'') {\em an} embedding, but they may not provide {\em the} embedding 
    (that we might have in mind). In fact, without a good definition of the desired embedding, 
    the embedding is not unique. %, so there is no {\em the} parametrization.
    
    \item Sometimes, the effects can be controlled if there is knowledge of the structure of the measurement function (e.g., Lipschitz constant). However, nuances in definitions of the output of algorithms, the increased complexity of algorithms, and the practice of layering algorithms on top of each other may make it much more difficult to control such effects. 
    In some special cases, additional measurements may allow one to reverse the 
    effect~\cite{singer_non-linear_2008,talmon_empirical_2013,peterfreund_local_2020,schwartz_intrinsic_2019,bertalan_transformations_2021}.

    \item In many (but not all) applications, the inferred manifold may reveal enough about the {\em topology} of the problem, or the distortion in the metric might be sufficiently small to be a sufficiently good proxy for the geometry. What is ``sufficiently good'' may depend on the downstream task.
    For example, the low-dimensional manifold {\em may} be a starting point for an analysis by an expert, regression or careful clustering, suspected outliers detection, and even for identification of clear modes. It {\em may} not be as 
    helpful for aligning data collected using different modalities (or even different algorithms applied to the same data) with different distortions 
    (see Appendix~\ref{sec:two cameras}), or for certain analyses of free energy associated with the distribution.

\end{itemize}

\section{Acknowledgments}

This work was supported by the grants NIH/R01GM136780 and AFOSR/FA9550-21-1-0317,
and by the Simons Foundation.

\printbibliography

%\newpage

\appendix

\subsection{Additional experiments}

This appendix shows several plots and numerical experiments that further
illustrate the discrepancy between the phenomenon manifold and the measurement
manifold.

\subsubsection{Manifold learning on the phenomenon manifold}
\label{sec:ml on x}

\begin{figure}[h!]
    \centering
    \begin{subfigure}{0.25\textwidth}
        \centering
        \includegraphics[width=\linewidth]{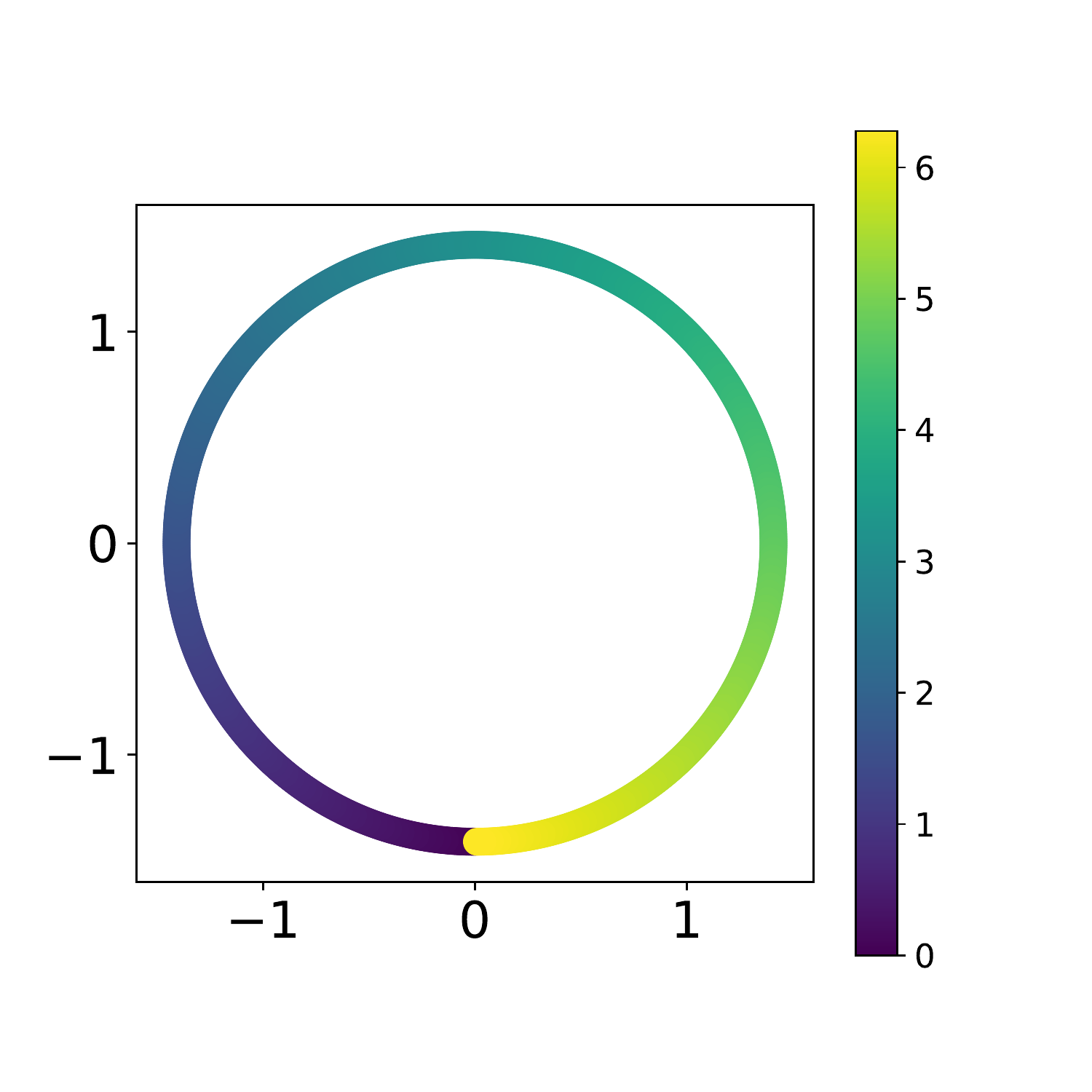}
        \caption{True angle}
        %\label{fig:sub1}
    \end{subfigure}%
    \begin{subfigure}{0.25\textwidth}
        \centering
        \includegraphics[width=\linewidth]{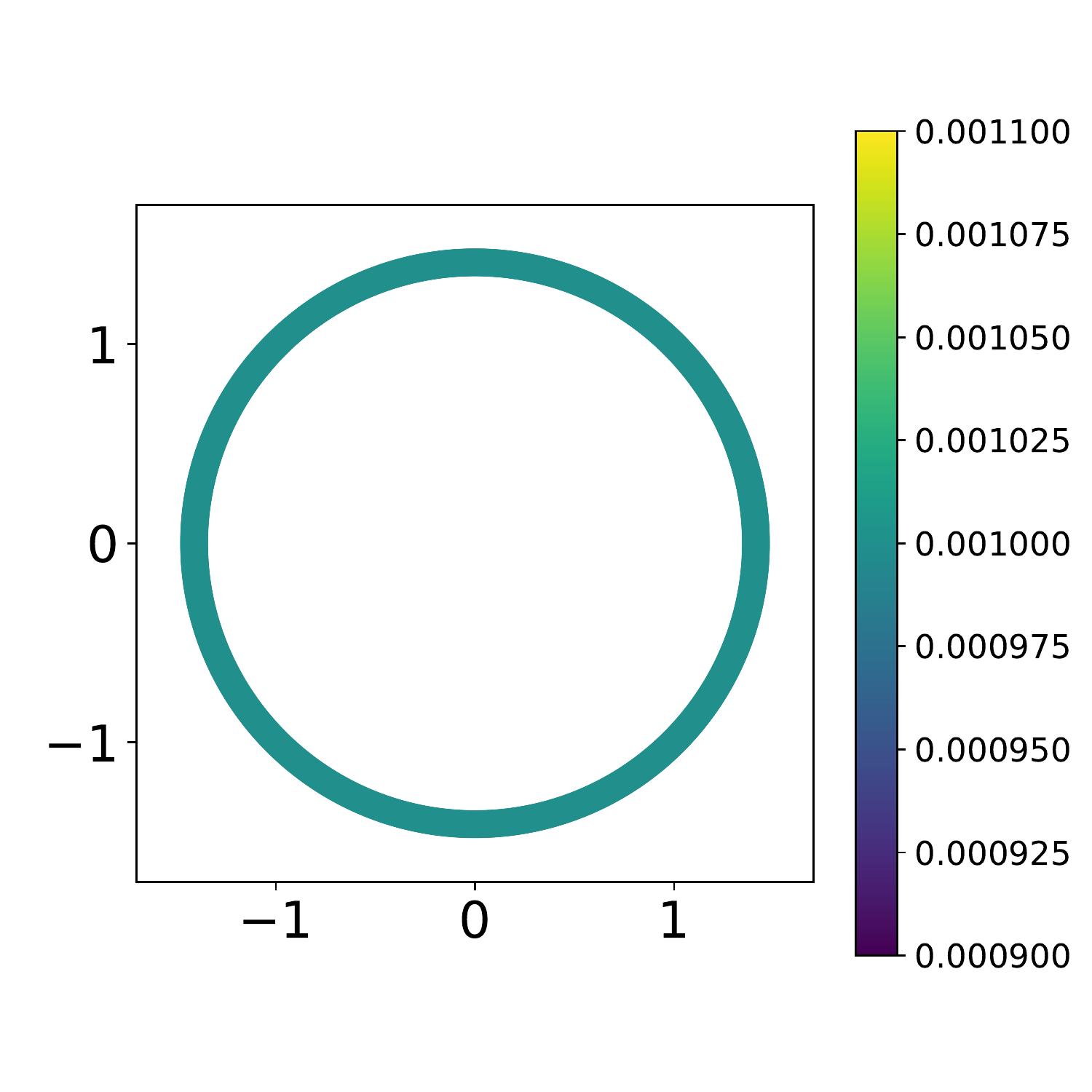}
        \caption{Local density}
        %\label{fig:sub1}
    \end{subfigure}
    \caption{The embedding resulting from applying diffusion maps to the 
    angles directly, represented as complex numbers of magnitude one.
        The coloring is given by the true angle in panel (a) and the local 
        density with $r=0.05$ in panel (b).}
    \label{fig:angles embedding}
\end{figure}

We have shown in the main text of the paper how applying a manifold learning
algorithm to the points on $\Yc$ leads to modes in the data, while
we know that the samples are uniformly distributed on the phenomenon 
manifold $\Xc$ (the orientation angles of the horse). 
In this section, we present the same manifold learning algorithm applied 
directly to the points in $\Xc$ (which are not available to us as measurements in our original problem setup) as a benchmark for comparison to the results in the main text, where we apply the algorithm to the measurements. 
In~Figure~\ref{fig:angles embedding}, we ran the diffusion maps algorithm to
the set of angles that generated the images, where we represented each
angle by a complex number with the magnitude of one and the given angle, and 
we used the Euclidean distance between them as the distance on $\Xc$.
As expected, the embedded points have constant local density (i.e., they 
are uniformly distributed).

\subsubsection{Local density of the measured data points}
\label{sec:density on y}

\begin{figure}[h!]
    \centering
    \begin{subfigure}{0.25\textwidth}
        \centering
        \includegraphics[width=\linewidth]{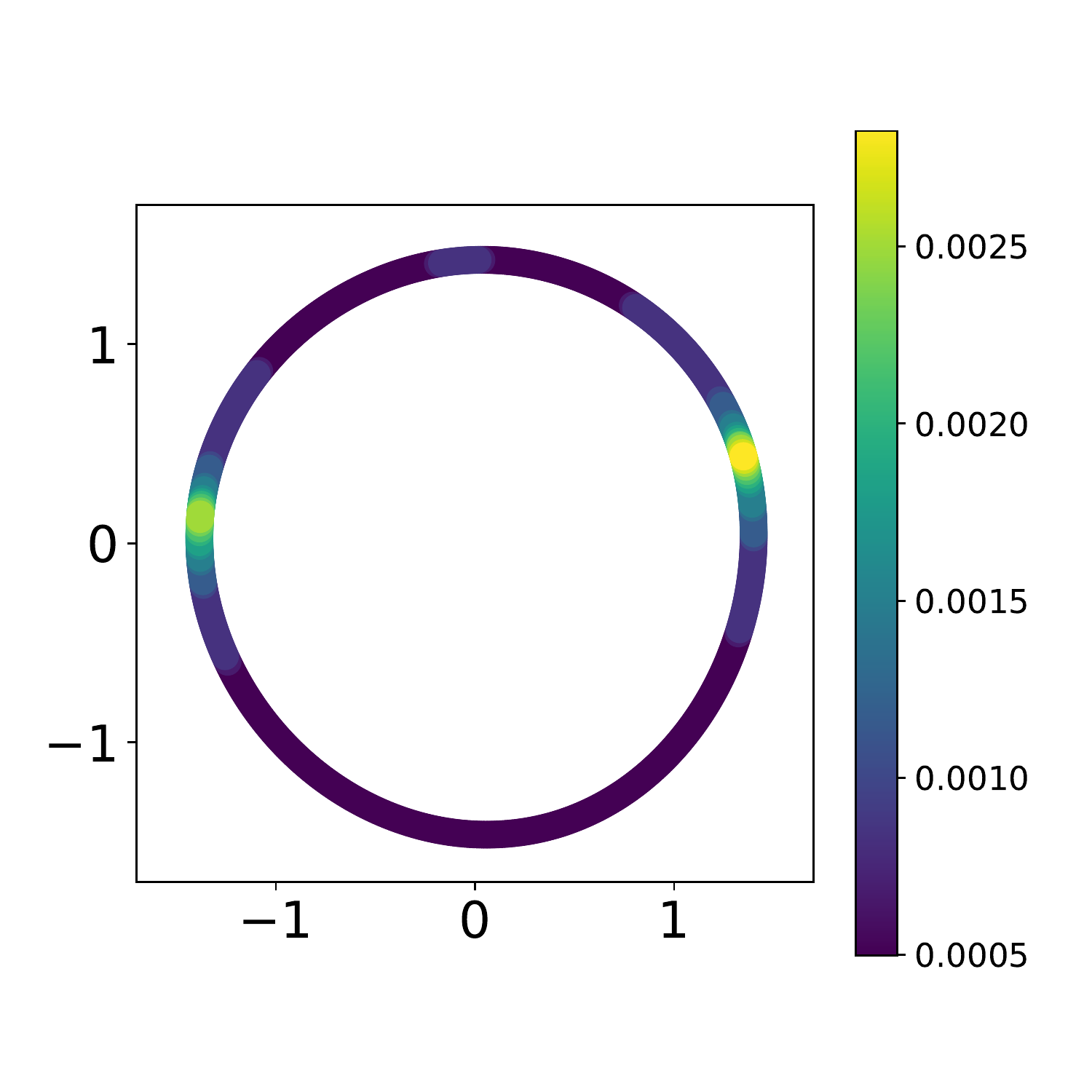}
        \caption{Local density on $\Yc$}
        %\label{fig:sub1}
    \end{subfigure}%
    \begin{subfigure}{0.25\textwidth}
        \centering
        \includegraphics[width=\linewidth]{embed_density_simulated_left_r0.05-eps-converted-to.pdf}
        \caption{Local density on $\Zc$}
        %\label{fig:sub1}
    \end{subfigure}
    \caption{Local density of the points on $\Yc$ in (a) and on $\Zc$ in (b). 
        The coordinates of the points are given by the embedding in $\Zc$ in both plots. 
        The radius for computing the local density on $\Yc$, $r=9$, was chosen 
        so that the maximum value of the density matches the maximum value of 
        the density on $\Zc$.}
    \label{fig:density on Y}
\end{figure}

The central claim of this paper is that applying manifold learning to measured
data without careful attention to the metric on $\Yc$ can lead to wrong conclusions about certain questions.
The point where metric distortions due to the measurement function are introduced 
in the processing pipeline is when computing pair-wise distances between the points on $\Yc$.
As a further check that this is indeed 
the case, in~Figure~\ref{fig:density on Y} we show the local density of the
images themselves (the points on $\Yc$): the coordinates of each image are
its embedding coordinates (the same as in the main text), and the coloring is given by the local density of
the images in panel (a) and the local density of the embeddings in panel (b).
The radius $r$ of the ball used to approximate the local density in (a) was chosen 
so that the maximum value of the local density is approximately the same as the 
maximum value of the local density in (b). Figure~\ref{fig:density on Y} shows that 
the same modes seen in the embedding are also present in the measured data.
The specific value may not be numerically identical due to subtleties in 
the definition of the embedding and the nature of the approximation. 
However, the same kind of effect is clearly visible.

\subsubsection{Two modalities and non-uniqueness}
\label{sec:two cameras}

\begin{figure}[h]
    \centering
    \includegraphics[width=0.8\linewidth]{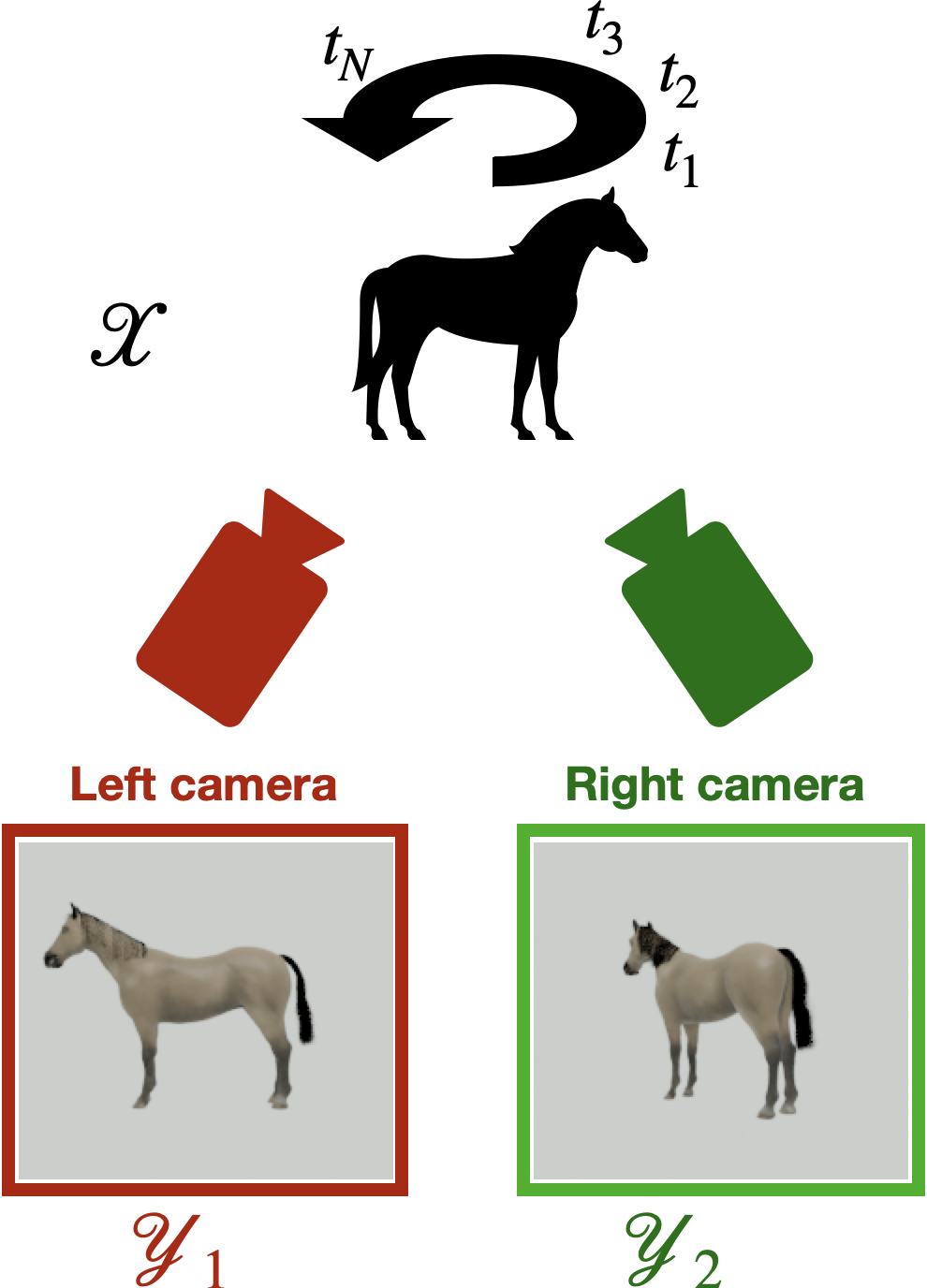}
    \caption{Setup of the numerical experiment with two measured datasets. 
        Using two cameras at different locations, we collect two distinct 
        sets of images on the measurement manifolds $\Yc_1$ and $\Yc_2$, respectively.}
    \label{fig:horse two cameras}
\end{figure}

To further illustrate the problem raised in this article, we now
consider a slightly modified setup to demonstrate the non-uniqueness of the parametrization. Instead of one camera, we take photographs
of the spinning rigid object using two different cameras placed at two distinct 
locations. We denote the two measurement functions by $f_1$ and $f_2$ and the two 
datasets by $\{y_i^1\}_{i=1}^n \subset \Yc_1$ and $\{y_i^2\}_{i=1}^n \subset \Yc_2$.
The images $y_i^1$ and $y_i^2$ at index $i$ taken by the two cameras respectively 
correspond to the same ground truth point $x_i$. This setup is illustrated 
in Figure~\ref{fig:horse two cameras}. The ambient dimension is $D=108000$, the 
number of images taken by each camera is $n=1000$, and the measurements are taken
at equal time intervals, so the distribution of the data on the phenomenon 
manifold $\Xc$ is uniform.

\begin{figure}[h]
    \centering
    \begin{subfigure}{0.25\textwidth}
        \centering
        \includegraphics[width=\linewidth]{embed_simulated_left-eps-converted-to.pdf}
        \caption{Left camera, true angle}
        %\label{fig:sub1}
    \end{subfigure}%
    \begin{subfigure}{0.25\textwidth}
        \centering
        \includegraphics[width=\linewidth]{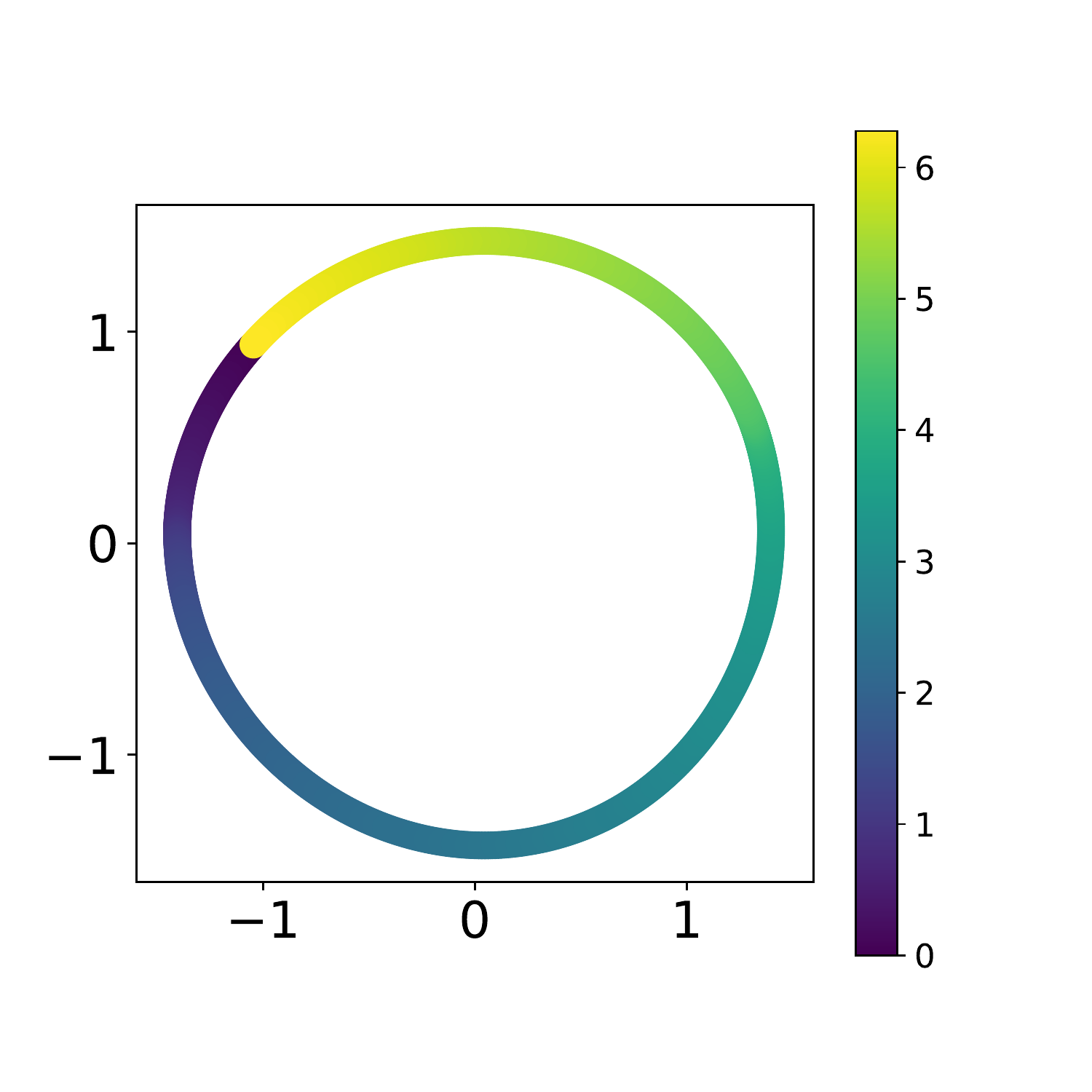}
        \caption{Right camera, true angle} 
        %\label{fig:sub2}
    \end{subfigure}
    \begin{subfigure}{0.25\textwidth}
        \centering
        \includegraphics[width=\linewidth]{embed_density_simulated_left_r0.05-eps-converted-to.pdf}
        \caption{Left camera, local density}
    \end{subfigure}%
    \begin{subfigure}{0.25\textwidth}
        \centering
        \includegraphics[width=\linewidth]{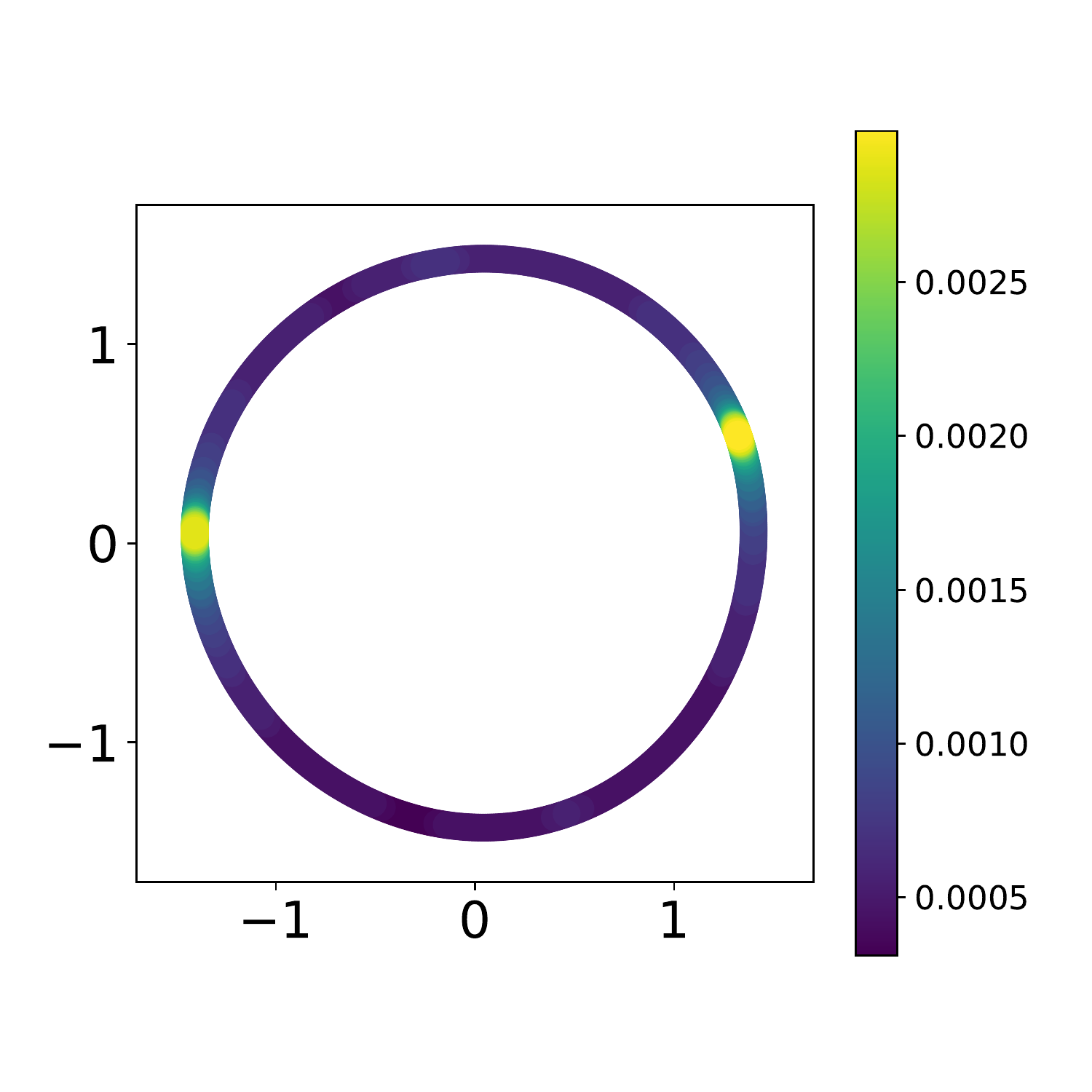}
        \caption{Right camera, local density}
    \end{subfigure}
    \caption{Embeddings obtained using images from the left 
        camera (left column) and the 
        right camera (right column). The coloring is given by 
        the true angle in the top row and the local density 
        of points ($r=0.05$) in the bottom row.}
    \label{fig:simulated results}
\end{figure}

\begin{figure}%[h]
    \centering
    \includegraphics[width=\linewidth]{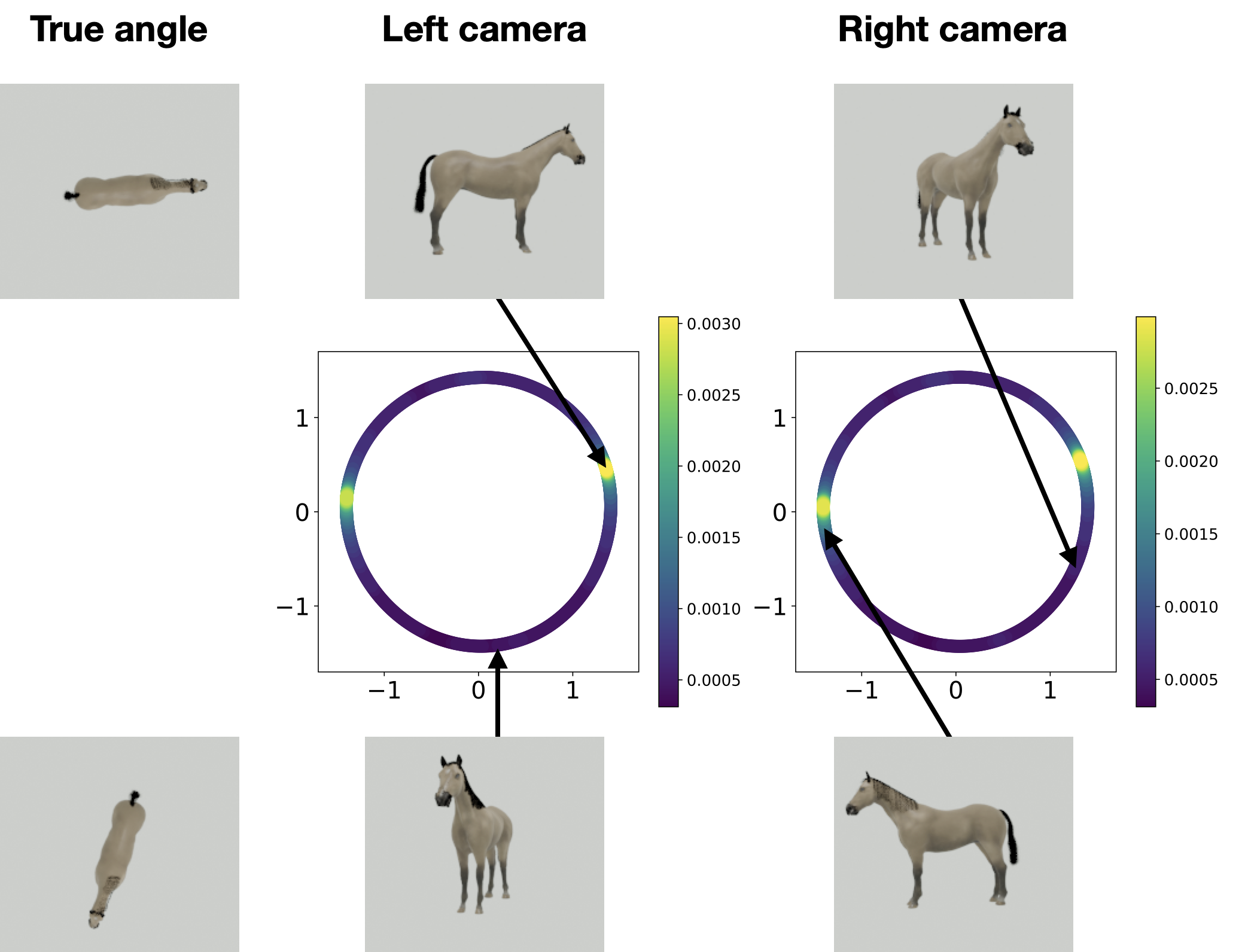}
    \caption{The embeddings with example images from the estimated distributions.
        The images in the top row correspond to a point in a 
        high-density region of the left camera embedding, and the images in the 
        bottom row correspond to a point in a low-density region
        of the left camera embedding. In each row of images, we show the top view
        of the object displaying the true orientation angle (left), the object 
        as seen by the left camera (middle), and the object as seen by the right 
        camera (right).}
    \label{fig:simulated examples}
\end{figure}

We apply the diffusion maps algorithm separately to each dataset, corresponding 
to samples from the two measurement manifolds $\Yc_1$ and $\Yc_2$. 
The resulting two-dimensional embeddings are shown in Figure~\ref{fig:simulated results}: the 
left column shows the embedding $\{\rho_1(y_i^1)\}_{i=1}^n$ obtained from the measurements
$\{y_i^1\}_{i=1}^n \subset \Yc_1$ and the right column shows the embedding 
$\{\rho_2(y_i^2)\}_{i=1}^n$ obtained from $\{y_i^2\}_{i=1}^n \subset \Yc_2$. 
In each panel, we show a scatter plot using the first two embedding coordinates 
given by the diffusion maps algorithm. 
We denote by $\Zc_1$ and $\Zc_2$ the resulting low-dimensional manifolds. Similarly 
to the one-camera experiment described in the main text, the one-dimensional torus topology of 
the orientation angles of the object is correctly identified (panels (a) and (b)). 
However, panels (c) and (d) show that the metric is distorted. In particular, the 
distributions of the images in both datasets are incorrectly shown to have two modes (we know that the true distribution is uniform). 
Moreover, the modes are not compatible: the mode observed in one camera does not correspond to the same true angles as those observed in the other camera.
%the embeddings obtained from the two datasets have distinct modes with no clear measure of which one is more accurate. 
This can be seen in Figure~\ref{fig:simulated examples}, where images from 
high and low-density regions on $\Zc_1$
correspond to seemingly arbitrary points on $\Zc_2$, whose high and
low-density regions correspond to different orientations on $\Xc$. This is, of
course, not surprising, given that each camera takes the photographs from 
different directions and considering the symmetries in this problem.

In this paper, we showed how attempts to solve a simple problem, such as 
identifying the dominant state of a system, can lead to incorrect answers when 
applying manifold learning to the measured data. While the experiment presented 
in the main text of the article shows that the answer we obtain can be incorrect
in a non-obvious way, the two-camera experiment presented in this appendix 
demonstrates that different measurement modalities can produce {\em different} 
answers, with no way to compare the quality of the two answers objectively.
More generally, this shows that the estimated measurement manifold is not
unique and not invariant to the measurement modality.

This experiment corresponds to several relatively common real-life settings.
One example is when multiple different algorithms are applied to the same data, 
perhaps with different processing pipelines. The processing pipelines 
are analogous to our different cameras and may not produce the same result.

Another real-life setting is when attempting to align measurements of an underlying 
phenomenon, acquired using two distinct modalities, calibrated differently and possibly taken 
on different days. This is the case, for example, when the data is collected in two 
separate batches, potentially in different laboratories. In this case, even if the 
two datasets are assumed to have the same distribution, batch effects are present due 
to potentially different experimental settings. Our experiments show that a rigid 
transformation cannot align embeddings obtained from such datasets.

\subsubsection{Top view measurements}
\label{sec:top view}

\begin{figure}%[h]
    \centering
    \begin{subfigure}{0.25\textwidth}
        \centering
        \includegraphics[width=\linewidth]{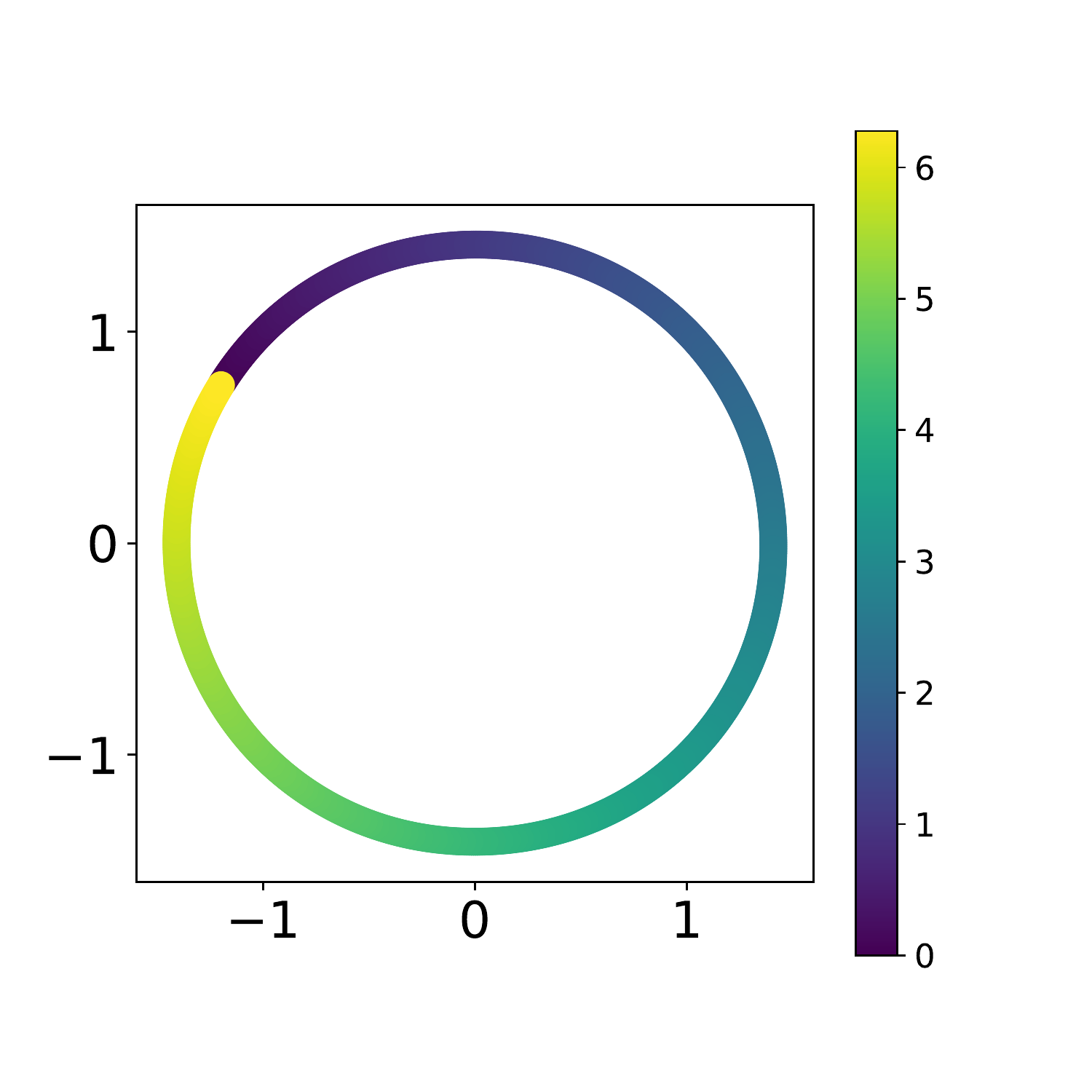}
        \caption{True angle}
        %\label{fig:sub1}
    \end{subfigure}%
    \begin{subfigure}{0.25\textwidth}
        \centering
        \includegraphics[width=\linewidth]{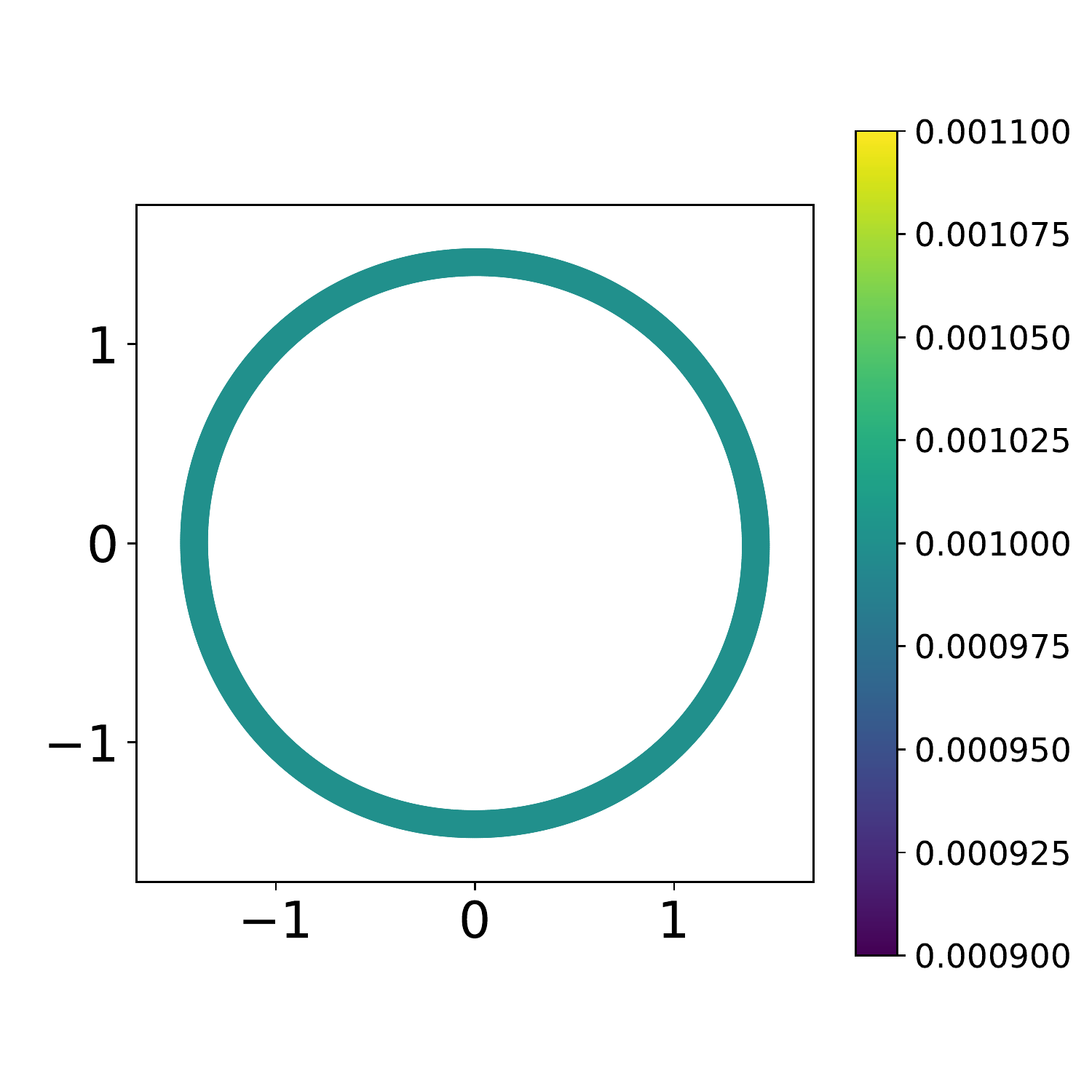}
        \caption{Local density}
        %\label{fig:sub1}
    \end{subfigure}
    \caption{Low-dimensional embedding of the images of the spinning horse 
        viewed from the top.
        The coloring is given by the true orientation angle of the horse in
        panel (a) and the local density of points ($r=0.05$) in panel (b).}
    \label{fig:main result top}
\end{figure}

In the previous experiment, we showed that different measurement functions lead to 
embeddings where the metric is distorted in different ways without knowing 
which one is more accurate. To further strengthen this argument, we show an 
example where the measurement function distorts the distances but preserves the local geometry.
The images of the rotating horse are captured from the top, and the resulting
embedding and density are shown in Figure~\ref{fig:main result top}. 
In practice, there is no way of knowing, only from the data, that we are in a case 
where the local geometry on $\Xc$ is preserved.
This experiment and the one in Appendix~\ref{sec:two cameras} reinforce 
that the embedding we obtain depends heavily on how the measurements are taken 
and that the solution we expect to see is not objectively better than other 
possible solutions, given the data.

\newpage
\vfill\null
%\newpage
\subsection{Diffusion maps}
\label{sec:dm}

For completeness, we present the diffusion maps algorithm used throughout the article, 
as described in~\cite{lafon_diffusion_2004} and adapted in~\cite{lederman_learning_2018}. 
The phenomena discussed in this paper are not specific to diffusion maps. Broadly 
interpreted, these issues appear in many machine learning and manifold learning problems 
in one form or another, except for special cases where they can be corrected or ``defined away.''

%\vspace{0.5cm}
%\begin{minipage}[]{0.465\textwidth}
    %\textbf{Diffusion maps algorithm}
    \vspace{0.1cm}\hrule\vspace{0.1cm}
    \textbf{Input}: Samples $\{y_i\}_{i=1}^n \subset \Yc$, a metric $d(y_i,y_j)$
    on $\Yc$, kernel width $\sigma$.

    \textbf{Output}: Low-dimensional embedding coordinates 
    $\{z_i\}_{i=1}^n \subset \mathbb{R}^s$ 
    and diffusion distances $\tilde{d}(z_i,z_j)$.

    \vspace{0.1cm}\hrule\vspace{0.1cm}
   
    \begin{enumerate}
        \item Compute the similarity matrix $W$:
            $$ W_{ij} = \exp\left(-\frac{d(y_i,y_j)^2}{\sigma} \right),
            \forall i,j=1,\ldots,n.$$

        \item Compute the diagonal normalization matrix 

            $$Q_{ii} = (\sum_{j=1}^n W_{ij})^{-1}.$$

        \item Normalize the kernel $\tilde{K}= QWQ$.

        \item Compute the second diagonal normalization matrix

            $$\tilde{Q}_{ii} = (\sum_{j=1}^n \tilde{K}_{ij})^{-1/2}.$$

        \item Normalize the kernel $K = \tilde{Q}\tilde{K}\tilde{Q}$.

        \item Compute the $n$ eigenvectors $u_0, u_1, \ldots, u_{n-1}$ 
            and eigenvalues $\lambda_0, \lambda_1, \ldots, \lambda_{n-1}$ of $K$.

        \item The $s$-dimensional embedding of each point $y_i$ is
            $$ z_i = \left(
                \lambda_1 \frac{u_1(i)}{u_0(i)},
                \lambda_2 \frac{u_2(i)}{u_0(i)},
                \ldots, 
                \lambda_s \frac{u_s(i)}{u_0(i)}
            \right)^T. $$
        \item The diffusion distance is 
            $\tilde{d}(z_i, z_j) = \|z_i-z_j\|_2.$
    \end{enumerate}
    %\vspace{0.1cm}\hrule
    %\vspace{0.1cm}
%\end{minipage}

\end{document}